\documentclass[letterpaper, 10 pt, conference]{ieeeconf}  

\IEEEoverridecommandlockouts                              

\usepackage{cite}
\usepackage{amsmath,amssymb,amsfonts}
\usepackage{algorithmic}
\usepackage{graphicx}
\usepackage{textcomp}
\def\BibTeX{{\rm B\kern-.05em{\sc i\kern-.025em b}\kern-.08em
    T\kern-.1667em\lower.7ex\hbox{E}\kern-.125emX}}
\usepackage{dsfont}
\usepackage{paralist}
\usepackage{cite}
\usepackage{tabularx}
\usepackage{cancel}
\usepackage{bm}
\usepackage{url}
\usepackage{mathtools}
\usepackage{textcomp}
\usepackage{nccmath}
\usepackage{xcolor}
\usepackage{multirow}
\usepackage{hhline}
\usepackage{booktabs}
\usepackage{flushend}
\usepackage{hyperref}
\usepackage{microtype}
\usepackage[table]{xcolor}
\usepackage[hang,flushmargin]{footmisc}

\title{\LARGE \bf
LAD-Drive: Bridging Language and Trajectory with Action-Aware Diffusion Transformers
}

\author{Fabian Schmidt$^{1,2,*}$, Karol Fedurko$^{1,*}$, Markus Enzweiler$^{1}$, Abhinav Valada$^{2}$
\thanks{$^{*}$ Equal Contribution.}%
\thanks{$^{1}$ Institute for Intelligent Systems, Esslingen University of Applied Sciences, Germany.}%
\thanks{$^{2}$ Department of Computer Science, University of Freiburg, Germany.}%
\thanks{This work has been funded by the BMFTR under support code 13FH544KB2 (ADRIVE-GPT). The authors alone are responsible for the content of the paper. The authors thank the DACHS data analysis cluster, hosted at Esslingen University and co-funded by the MWK within the DFG's "Großgeräte der Länder" program, for providing the computational resources necessary for this research.}%
}

\begin{document}

\maketitle
\thispagestyle{empty}
\pagestyle{empty}

\begin{abstract}

While multimodal large language models (MLLMs) provide advanced reasoning for autonomous driving, translating their discrete semantic knowledge into continuous trajectories remains a fundamental challenge. 
Existing methods often rely on unimodal planning heads that inherently limit their ability to represent multimodal driving behavior. 
Furthermore, most generative approaches frequently condition on one-hot encoded actions, discarding the nuanced navigational uncertainty critical for complex scenarios. 
To resolve these limitations, we introduce LAD-Drive, a generative framework that structurally disentangles high-level intention from low-level spatial planning. 
LAD-Drive employs an action decoder to infer a probabilistic meta-action distribution, establishing an explicit belief state that preserves the nuanced intent typically lost by one-hot encodings. 
This distribution, fused with the vehicle's kinematic state, conditions an action-aware diffusion decoder that utilizes a truncated denoising process to refine learned motion anchors into safe, kinematically feasible trajectories.
Extensive evaluations on the LangAuto benchmark demonstrate that LAD-Drive achieves state-of-the-art results, outperforming competitive baselines by up to 59\% in Driving Score while significantly reducing route deviations and collisions.
We will publicly release the code and models on \url{https://github.com/iis-esslingen/lad-drive}.

\end{abstract}

\section{Introduction}

The field of end-to-end autonomous driving has shifted significantly with the integration of pretrained Large Language Models (LLMs). 
Early frameworks leverage the vast world knowledge and reasoning capabilities of LLMs to navigate complex, long-tail traffic scenarios. 
Specifically, pioneering approaches~\cite{mao2023gpt, xu2024drivegpt4, tiandrivevlm} reformulate continuous trajectory planning as an autoregressive text generation task, predicting spatial waypoints directly within the discrete language space. 
However, this direct translation suffers from a severe modality mismatch between discrete language tokens and continuous action spaces. 
The inherent numerical insensitivity of discrete textual tokenization not only frequently generates physically infeasible, malformed outputs that trigger downstream parsing failures~\cite{li2025recogdrive}, but also yields unstable predictions that severely degrade long-term planning~\cite{xie2026latentvla}.

To avoid these pitfalls, many approaches employ dedicated regression heads to map vision-language features directly to continuous trajectories~\cite{shao2024lmdrive, winter2025bevdriver, renz2025simlingo}. 
While computationally efficient, these coupled architectures face two critical limitations. 
Architecturally, forcing a single network head to simultaneously resolve high-level semantic intentions (what to do) and low-level spatial planning (how to execute it) creates a severe entanglement that can overwhelm model capacity. 
Mathematically, regression objectives (such as L1 or L2 loss) are fundamentally ill-equipped to handle the multimodal nature of driving. They inherently average across divergent but equally valid behaviors (e.g., turning left versus right to avoid an obstacle). 
This mode averaging can yield to physically infeasible predictions, leading to poor generalization in complex scenarios~\cite{yin2025diffrefiner}.

\begin{figure}[t!]
\centering
\includegraphics[width=1.0\linewidth]{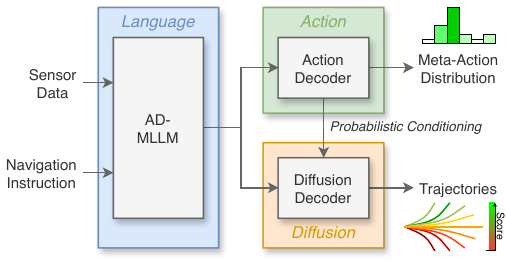}
\caption{LAD-Drive disentangles high-level semantic reasoning from low-level spatial planning across three modules. First, the \textit{Language} module uses a multimodal LLM (AD-MLLM) to synthesize sensor data and navigation instructions into contextualized hidden states. Second, the \textit{Action} decoder infers a discrete meta-action distribution directly from these states. Finally, the \textit{Diffusion} decoder utilizes this probabilistic conditioning to iteratively refine noisy priors into kinematically consistent, multimodal trajectories.}
\label{fig:architecture_overview}
 \vspace{-2mm}
\end{figure}

To capture multimodal driving behaviors, recent literature has explored two primary architectural paradigms~\cite{oksuz2025foundation}. 
The first reframes trajectory prediction as a classification task by discretizing the continuous action space into a finite vocabulary of specialized action tokens~\cite{chen2024vadv2, zhou2025autovla}. 
While this prevents mode averaging by forcing the model to output a categorical distribution, it inherently introduces quantization errors that compromise smooth low-level planning~\cite{xie2026latentvla}. 
The second paradigm shifts toward generative modeling. By leveraging diffusion models to explicitly learn the underlying distribution of future motion, this approach allows flexible trajectory prediction and effectively captures the complex multimodality of human driving~\cite{ yin2025diffrefiner}.

While diffusion models excel at generating diverse and kinematically plausible paths~\cite{wei2025parkdiffusion, wei2026parkdiffusion++}, effectively conditioning these planners on the semantic reasoning of an LLM remains a critical bottleneck. 
As our analysis reveals, feeding raw LLM hidden states directly into a planner introduces semantic noise and causal confusion. 
Conversely, attempts to simplify this conditioning introduce new bottlenecks. Rigid, one-hot encoded actions~\cite{zhang2024ad, jiang2024senna, jiang2025diffvla} default to a winner-takes-all paradigm that discards critical uncertainty, while opaque implicit planning tokens~\cite{fu2025orion} obscure the agent's semantic intent, making it impossible to explicitly model or interpret the underlying probabilistic belief state.

To bridge this intention-action gap, we introduce LAD-Drive (\textbf{L}anguage and \textbf{A}ction-Aware \textbf{D}iffusion \textbf{Drive}), an end-to-end framework that structurally disentangles semantic intention from geometric execution. As shown in Fig.~\ref{fig:architecture_overview}, LAD-Drive employs an \textit{action decoder} to infer a probabilistic meta-action distribution, thereby forming an explicit belief state that preserves critical uncertainty. 
To execute this intention without causal confusion, an action-aware \textit{diffusion decoder} grounds the generation process in both this belief state and distilled LLM features. 
By utilizing clustered anchors as trajectory priors within a truncated diffusion schedule, LAD-Drive efficiently refines safe, kinematically feasible paths in just two denoising steps.

In summary, our core contributions are the structural disentanglement of semantic reasoning and spatial planning through explicit probabilistic conditioning, and the design of a highly efficient action-aware diffusion decoder that produces multimodal trajectories. 
Extensive evaluations and systematic ablation studies validate this paradigm, confirming the necessity of our dual-conditioning strategy and demonstrating that LAD-Drive achieves state-of-the-art performance on the LangAuto benchmark~\cite{shao2024lmdrive}. 
We make the code and trained models publicly available on \url{https://github.com/iis-esslingen/lad-drive}.

\section{Related Work}

\subsection{End-to-End Autonomous Driving}
The paradigm of end-to-end autonomous driving has evolved significantly from simple regression baselines to sophisticated generative frameworks.
UniAD~\cite{hu2023planning} pioneered the unified approach by integrating perception, prediction, and planning into a query-based transformer architecture, enabling effective multi-task interaction.
To better address the inherent uncertainty in driving behaviors, VADv2~\cite{chen2024vadv2} introduced probabilistic planning, which discretizes the continuous action space into a trajectory vocabulary to explicitly model the distribution of future actions.

Building on the need for continuous and multimodal output, recent research has increasingly leveraged diffusion models, with a specific focus on optimizing inference efficiency and guidance~\cite{zheng2025diffusionplanner, liao2025diffusiondrive, xing2025goalflow, yin2025diffrefiner, liu2025bridgedrive}. 
DiffusionPlanner~\cite{zheng2025diffusionplanner} formulates planning as a conditional generative task, utilizing a diffusion transformer to jointly model the ego-vehicle and surrounding agents with classifier guidance. 
Addressing the inference latency typical of such models, DiffusionDrive~\cite{liao2025diffusiondrive} proposes a truncated diffusion policy that initializes the denoising process from learned anchors rather than random noise. 
GoalFlow~\cite{xing2025goalflow} further optimizes this generation by introducing goal-driven flow matching, conditioning the trajectory on estimated goal points to ensure consistency and efficiency.
Adopting a coarse-to-fine paradigm, DiffRefiner~\cite{yin2025diffrefiner} employs a diffusion refiner to iteratively improve initial trajectory proposals by enforcing fine-grained semantic consistency. 
Most recently, BridgeDrive~\cite{liu2025bridgedrive} challenges the theoretical asymmetry of prior anchor-based methods by formulating planning as a symmetric diffusion bridge process that connects the current state to a target state to generate physically rigorous trajectories.\looseness=-1

\subsection{LLM-based Autonomous Driving}

The integration of LLMs into autonomous driving has progressed from high-level reasoning in simplified environments~\cite{fu2024humandrive} to closed-loop trajectory planning in complex simulations~\cite{shao2024lmdrive, winter2025bevdriver, renz2025simlingo, zhang2025adadrive, zhang2025vldrive, kim2026storm, zhang2024ad, jiang2024senna, schmidt2025enhancing}. 
This shift to realistic, closed-loop control was pioneered by LMDrive~\cite{shao2024lmdrive}, which integrated multimodal sensor data with natural language instructions. Subsequent works have focused on optimizing this foundation. 
Building on this architecture, BEVDriver~\cite{winter2025bevdriver} enhances the perception module to generate more comprehensive BEV representations, thereby optimizing both spatial awareness and semantic alignment. 
SimLingo~\cite{renz2025simlingo} targets language-action alignment within a vision-only framework to ensure consistency between linguistic reasoning and executed maneuvers. 
Addressing computational efficiency, AdaDrive~\cite{zhang2025adadrive} implements an adaptive slow-fast system to gate LLM activations, whereas VLDrive~\cite{zhang2025vldrive} reduces overhead by using a lightweight LLM and dynamic visual pruning.
Similarly, SToRM~\cite{kim2026storm} mitigates the high computational overhead of MLLMs by introducing a supervised token reduction mechanism, thereby improving driving performance.
To bridge the gap between abstract reasoning and precise planning, hierarchical frameworks such as AD-H~\cite{zhang2024ad} and Senna~\cite{jiang2024senna} employ an MLLM to infer discrete, high-level meta-actions that subsequently condition a dedicated low-level planner for discrete trajectory prediction.

To overcome the unimodal limitations of regression-based planners, recent research has shifted toward generative models that can capture multimodal trajectory distributions~\cite{fu2025orion, jiang2025diffvla, li2025recogdrive, ma2025dvlm, li2026sgdrive, liu2025drivepi}.
A primary challenge lies in effectively conditioning these models.
ORION~\cite{fu2025orion} employs an implicit planning token within an LLM to guide a generative planner, while DiffVLA~\cite{jiang2025diffvla} utilizes an MLLM to produce one-hot encoded lateral and longitudinal commands for a hybrid sparse-dense diffusion planner.
Taking a hierarchical approach, SGDrive~\cite{li2026sgdrive} forecasts world knowledge in a scene-agent-goal manner to serve as a latent condition.
To overcome the sequential restrictions of autoregressive models, dVLM-AD~\cite{ma2025dvlm} employs a discrete diffusion MLLM with bidirectional attention, leveraging iterative denoising to enhance controllability and consistency between high-level reasoning and low-level planning.
Finally, ensuring that generated trajectories are physically grounded and safe remains critical.
DrivePI~\cite{liu2025drivepi} incorporates 3D occupancy and flow as auxiliary tasks for spatial-temporal grounding, whereas ReCogDrive~\cite{li2025recogdrive} integrates Diffusion Group Relative Policy Optimization to enforce safety and comfort constraints.

\begin{figure*}[t!]
\centering
\includegraphics[width=\linewidth]{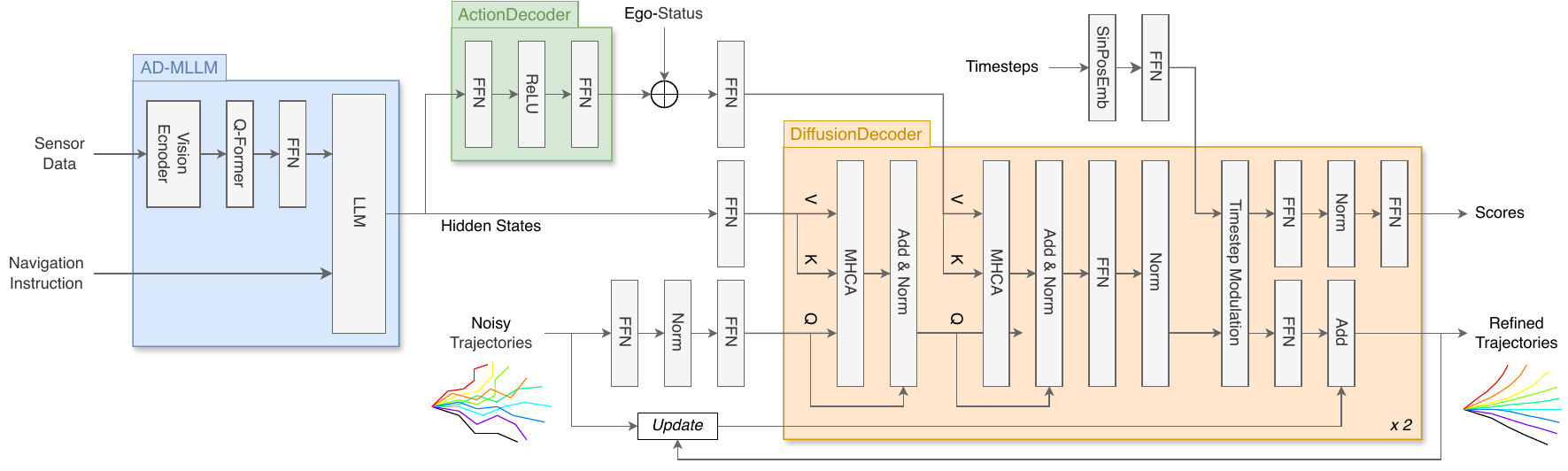}
\caption{Detailed architecture of LAD-Drive. The framework utilizes an AD-MLLM backbone to process multi-view images, LiDAR point clouds, and navigation instructions into contextualized hidden states. A dedicated action decoder (green) infers a meta-action distribution from these states, which is fused with the vehicle's ego-status to form a joint state-intent representation. The diffusion decoder (orange) employs a shared transformer-based block to iteratively refine noisy trajectory anchors over two denoising steps, with each step's output updating the input for the next. This refinement process utilizes sequential multi-head cross-attention (MHCA) layers to ground trajectory generation in both the global visual-language context and the predicted probabilistic intent, outputting refined trajectories and their corresponding confidence scores.}
\label{fig:architecture_detailed}
\end{figure*}

In contrast to existing approaches that rely on opaque implicit conditioning~\cite{fu2025orion} or rigid one-hot encodings~\cite{zhang2024ad, jiang2024senna, jiang2025diffvla}, LAD-Drive structurally disentangles high-level intention from low-level planning by employing a dedicated action decoder to predict an explicit meta-action distribution. This probabilistic conditioning enables the system to navigate high-uncertainty scenarios effectively, guiding a continuous diffusion decoder to generate robust trajectories that strictly adhere to semantic navigation instructions.

\section{Method}

We formulate the task of end-to-end autonomous driving as learning a conditional distribution $P(\tau \mid \mathcal{O}, \mathcal{I})$ that maps multimodal sensor observations $\mathcal{O}$ and natural language navigation instructions $\mathcal{I}$ to a future trajectory $\tau = \{w_1, \dots, w_T\}$, where $w_i \in \mathbb{R}^2$ represents waypoints in the vehicle's local coordinate frame. 
While regression approaches based on MLPs or GRUs approximate the conditional expectation $\mathbb{E}[\tau \mid \mathcal{O}, \mathcal{I}]$, this formulation can struggle in highly multimodal scenarios where multiple paths are valid by averaging divergent paths into suboptimal trajectories. 
Instead, we propose to explicitly preserve the multimodality of the driving scene.
We achieve this by structurally disentangling high-level intention from low-level execution by introducing an intermediate probabilistic belief state $\hat{\mathbf{p}} = P(\mathcal{A} \mid \mathcal{O}, \mathcal{I})$, representing the meta-action distribution over the discrete action space $\mathcal{A}$. 
The trajectory generation is then formulated as a diffusion process conditioned jointly on the observations and this explicit belief state:

\begin{equation}
    P(\tau \mid \mathcal{O}, \mathcal{I}) \approx P(\tau \mid \hat{\mathbf{p}}, \mathcal{O}, \mathcal{I})
\end{equation}

To instantiate this formulation, our model utilizes an MLLM tailored for autonomous driving (AD-MLLM) as its central semantic backbone, following the work of~\cite{shao2024lmdrive}. 
The architecture leverages an LLM to process the multimodal context, integrating both sensor and natural language inputs. Specifically, a Vision Encoder first transforms multi-view camera and LiDAR features into a unified Bird's-Eye-View (BEV) representation. 
Utilizing a Q-Former~\cite{li2023blip}, the model then compresses these high-dimensional BEV features into a fixed set of learned scene tokens that are projected into the language space.

By treating the scene tokens and navigational instructions as a causal sequence, the LLM predicts a contextually aware latent representation of hidden states. 
To form the explicit belief state, a dedicated action decoder infers the meta-action distribution directly from these states. 
These outputs, combined with the vehicle's ego-status, serve as probabilistic conditioning signals for a downstream diffusion decoder, that iteratively refines noisy trajectories into refined trajectories that are kinematically consistent and semantically grounded in the perceived scene. 
The architecture is shown in Fig.~\ref{fig:architecture_detailed}.

\subsection{Action Decoder}
The action decoder infers high-level driving actions from the LLM's contextualized hidden states $\mathbf{h}_{\text{LLM}} \in \mathbb{R}^{4096}$. 
By explicitly supervising this module, we enforce an instruction-action alignment that bridges the gap between semantic intent and geometric execution, thereby mitigating ambiguous or causally inconsistent maneuvers. 
Furthermore, outputting discrete, human-understandable actions significantly enhances the overall interpretability of the model's behavior. 
Following the lateral action space defined in~\cite{fu2025orion}, we define the discrete set of lateral maneuvers as $\mathcal{A}_{\text{lat}}$, comprising six categories: \textit{Lane Follow}, \textit{Straight}, \textit{Left}, \textit{Right}, \textit{Lane Change Left}, and \textit{Lane Change Right}. 
As illustrated in Fig.~\ref{fig:architecture_detailed}, the decoder consists of a multi-layer feed-forward neural network (FFN) that outputs a probability distribution $\hat{\mathbf{p}}_{\text{lat}} \in \mathbb{R}^{|\mathcal{A}_{\text{lat}}|}$ via a terminal softmax layer.

While our architecture is designed to support decoupled lateral and longitudinal heads, our empirical analysis indicates that conditioning the diffusion process solely on lateral intent yields superior performance. 
We hypothesize that because longitudinal dynamics are largely governed by the vehicle's ego-status and immediate spatial constraints, a high-level longitudinal head introduces redundant semantic noise. 
Consequently, for our primary configuration, the probabilistic belief state $\hat{\mathbf{p}}$ is formed exclusively from the lateral action distribution $\hat{\mathbf{p}}_{\text{lat}}$. 
This provides a focused conditioning signal that resolves navigational ambiguity without compromising low-level planning stability.

\subsection{Diffusion Decoder}
For low-level planning, we propose a generative diffusion decoder based on the truncated diffusion policy introduced in~\cite{liao2025diffusiondrive}. 
Unlike existing methods that leverage explicit spatial representations, such as BEV maps, our decoder operates directly on the contextualized hidden states $\mathbf{h}_{\text{LLM}}$. 
Because these embeddings encapsulate a rich, compressed representation of the driving scene, historical context and navigational instructions, the generative process can flexibly condition on this semantic knowledge to natively preserve the multimodality of the driving environment.

\begin{figure}
\centering
\includegraphics[width=0.9\linewidth]{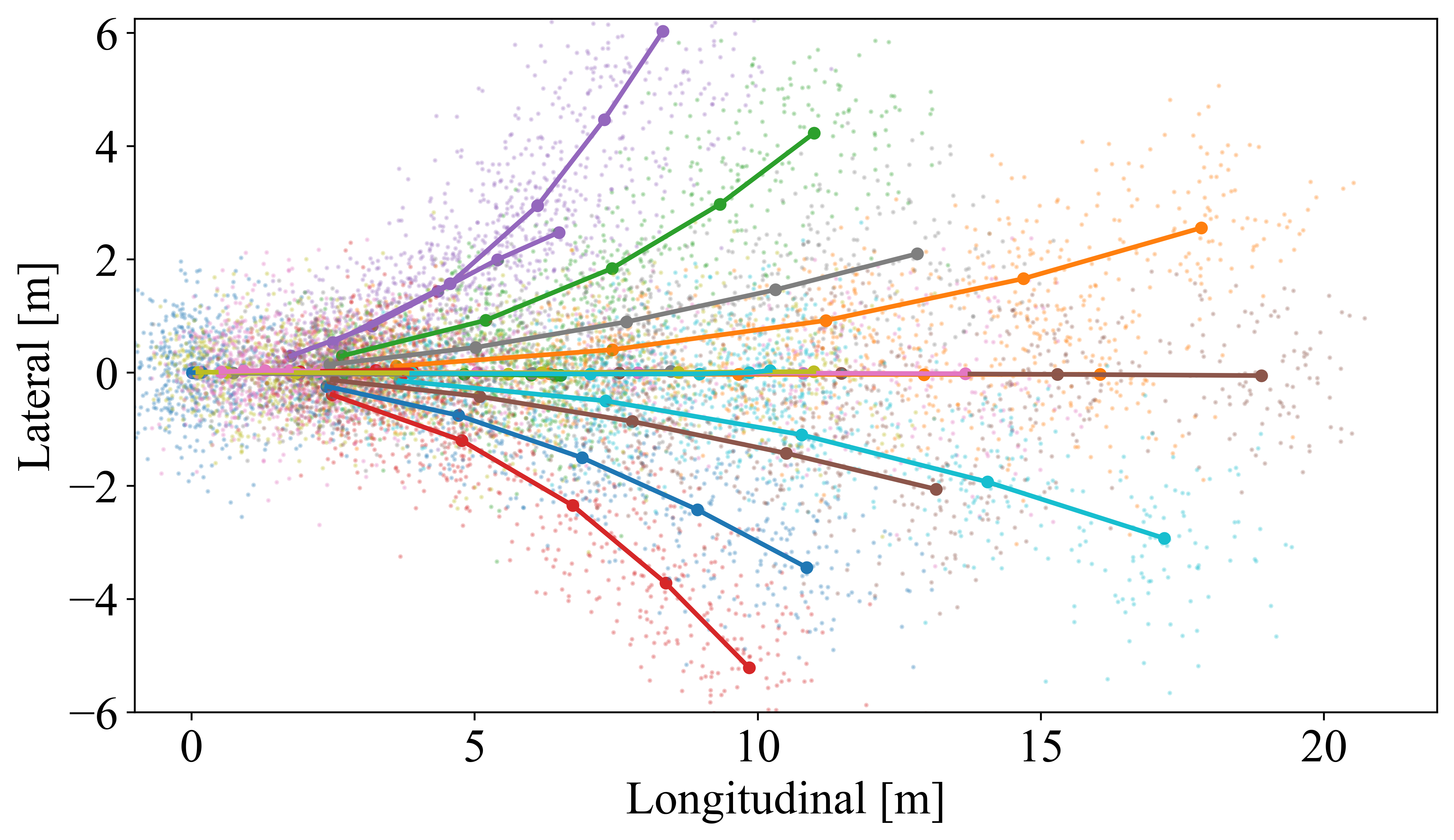}
\caption{Anchor trajectories generated via k-means clustering on training trajectories to provide motion priors for the initialization of the truncated diffusion process.}
\label{fig:anchors}
\vspace{-0.3cm}
\end{figure}

\paragraph{Anchor Encoding}
The diffusion decoder requires a set of anchor trajectories $\mathbf{A} = \{\mathbf{a}_i \in \mathbb{R}^{T \times 2}\}_{i=1}^{N_a}$, where $T=5$ waypoints and $N_a=20$ anchors. 
While previous works~\cite{liao2025diffusiondrive} utilize static anchors derived from datasets such as Navsim~\cite{dauner2024navsim}, we instead propose to domain-align our anchors to optimally capture the underlying properties of the operational domain.
Consequently, we generate this domain-aligned set of anchors by applying k-means clustering directly to the expert trajectories of our target training distribution, as shown in Fig.~\ref{fig:anchors}.
This data-driven initialization ensures that the motion priors accurately reflect the specific kinematic constraints and speed profiles of the evaluation environment, enabling the model to refine feasible paths with reduced planning complexity.
To embed these motion priors, a 2-layer MLP encoder maps them into a shared latent space, yielding $\mathbf{z}_a \in \mathbb{R}^{N_a \times d}$, where $d=512$.

\paragraph{Feature Bottleneck}
While recent methods such as ORION~\cite{fu2025orion} directly utilize raw, high-dimensional LLM hidden states as ``planning tokens'' to condition trajectory prediction, they report significant challenges in aligning these reasoning features with diffusion-based action spaces. 
Our empirical analysis confirms this. We observe that directly coupling the full 4096-dimensional LLM representation to the diffusion decoder leads to severely degraded driving performance (detailed in Sec.~\ref{sec:experiments}). 
We attribute this to a severe representation mismatch and dimensionality overload. As the raw LLM embeddings are heavily optimized for autoregressive linguistic prediction, they encapsulate broad, abstract world knowledge that is extraneous to low-level spatial planning. 
Injecting this high-dimensional semantic noise directly into the denoising process overwhelms the trajectory refinement.

To resolve this alignment issue and stabilize the generative process, we introduce a structural feature bottleneck that maps the LLM hidden states $\mathbf{h}_{\text{LLM}}$ into a compact $d$-dimensional latent space via a two-layer MLP, resulting in $\mathbf{h}'_{\text{LLM}} \in \mathbb{R}^{d}$. 
As shown in Fig.~\ref{fig:architecture_detailed}, this projection acts as a semantic filter based on the information bottleneck principle, ensuring the diffusion decoder processes only the most relevant features pertaining to navigational intent and spatial context. 

\paragraph{State-Intent Conditioning}
To ground the trajectory decoding, we introduce a joint state-intent representation. 
While we explicitly preserve navigational uncertainty by conditioning on the continuous probabilistic belief state $\hat{\mathbf{p}}$, high-level semantic intent alone is insufficient for low-level planning. 
To ensure the generated paths are kinematically feasible, the trajectory formulation must be strictly anchored to the vehicle's current physical dynamics. 
Therefore, as illustrated in Fig.~\ref{fig:architecture_detailed}, we concatenate the belief state $\hat{\mathbf{p}}$ with the vehicle's ego-status $\mathbf{e}$ (current velocity and yaw angle) to form a unified conditioning vector $\mathbf{c} = \mathbf{e} \oplus \hat{\mathbf{p}}$. 
This fusion explicitly couples \textit{what} the vehicle intends to do with \textit{how} it is currently moving, providing a comprehensive kinematic-semantic prior. 
Finally, this joint vector is mapped into the shared latent space via a two-layer MLP to obtain $\mathbf{z}_{\text{si}} \in \mathbb{R}^d$.

\paragraph{Iterative Refinement}
Having established the requisite geometric, semantic, and kinematic representations, the diffusion decoder begins the iterative generation process. 
Following the truncated diffusion paradigm~\cite{liao2025diffusiondrive}, the decoder performs $N=2$ refinement steps on k-means anchors initialized with Gaussian noise.
By starting from these noisy priors rather than pure noise, the progressive denoising process efficiently shifts the candidate trajectories toward scene-adaptive, kinematically feasible paths.

To ensure the generated trajectories are contextually aware, the anchor embeddings $\mathbf{z}_a$ first attend to the projected LLM hidden states $\mathbf{h}'_{\text{LLM}}$ via a multi-head cross-attention (MHCA) layer, formulated as $\mathbf{h}_t = \text{MHCA}(\mathbf{z}_a, \mathbf{h}'_{\text{LLM}}, \mathbf{h}'_{\text{LLM}})$. 
Functioning as a context attention mechanism, this block grounds the trajectory in the global visual-language context, enabling the planner to focus on the perceived scene layout and the provided navigation instructions.
This is followed by a second MHCA layer that integrates the state-intent representation $\mathbf{z}_{\text{si}}$: $\mathbf{h}_t = \text{MHCA}(\mathbf{h}_t, \mathbf{z}_{\text{si}}, \mathbf{z}_{\text{si}})$. 
Operating as a guidance attention mechanism, this step explicitly conditions the refinement process on the predicted intention and physical state. 
This ensures the generated path adheres to the high-level decision made by the action decoder while strictly respecting the vehicle's dynamic limits. 
By structuring the diffusion process with these sequential conditioning steps, the model effectively resolves the disconnect between high-level semantic planning and low-level execution.

The fused features $\mathbf{h}_t$ are subsequently processed by an FFN and modulated by a timestep embedding layer, which dynamically adjusts the latent features via scale and shift parameters based on the current diffusion step. 
Finally, the decoder branches into two specialized MLP heads: a waypoint head predicts offsets that iteratively deform the initial noisy anchors via a residual connection, and a classification head generates corresponding trajectory confidence scores.

\subsection{Training Strategy}
For training the diffusion decoder, we adapt the truncated denoising schedule from~\cite{liao2025diffusiondrive}. 
The diffusion decoder $f_{\text{dec}}$ processes the noisy anchors $\mathbf{Y}_{\text{noisy}}$ alongside our explicitly defined conditioning representations to predict $N_a$ refined trajectories $\mathbf{\hat{Y}} = \{\mathbf{\hat{y}}_i\}_{i=1}^{N_a}$ and their confidence scores $\mathbf{\hat{s}} = \{\hat{s}_i\}_{i=1}^{N_a}$:
\begin{equation}
    \mathbf{\hat{Y}}, \mathbf{\hat{s}} = f_{\text{dec}}(\mathbf{Y}_{\text{noisy}}, \mathbf{h}'_{\text{LLM}}, \mathbf{z}_{\text{si}}).
\end{equation}

To supervise the multimodality of the predictions without inducing mode collapse, we employ a closest-anchor matching strategy. 
The anchor with the minimum distance to the ground-truth trajectory $\mathbf{y}_{\text{GT}}$ is assigned the positive label $s_i = 1$, while all others are marked $s_i = 0$. 
The planning objective combines a binary cross-entropy (BCE) classification loss across all anchors with an L1 regression loss ($\mathcal{L}_{\text{wp}}$) applied exclusively to the positive anchor $\mathbf{\hat{y}}_{\text{pos}}$:
\begin{equation}
    \mathcal{L}_{\text{plan}} = \lambda_{\text{wp}} \mathcal{L}_{\text{wp}}(\mathbf{\hat{y}}_{\text{pos}}, \mathbf{y}_{\text{GT}}) + \lambda_{\text{cls}} \sum_{i=1}^{N_a} \text{BCE}(\hat{s}_i, s_i), 
\end{equation}
where $\lambda_{\text{wp}} = 8$ and $\lambda_{\text{cls}} = 10$ are hyperparameters balancing the two objectives.

The final training objective incorporates a standard cross-entropy (CE) loss to explicitly supervise the action decoder. 
Using the one-hot encoded ground-truth action $\mathbf{a}_{\text{GT}}$, the total loss is formulated as:
\begin{equation}
    \mathcal{L}_{\text{total}} = \mathcal{L}_{\text{plan}} + \lambda_{\text{action}}\text{CE}(\hat{\mathbf{p}}, \mathbf{a}_{\text{GT}}).
\end{equation}

We adopt a two-stage training strategy designed to decouple the learning of spatial motion from semantic reasoning.

\paragraph{Spatial Grounding} 
During the initial stage, the loss contribution of the action decoder is masked ($\lambda_{\text{action}} = 0$). 
The primary objective is to train the Q-Former~\cite{li2023blip} and the diffusion decoder to generate physically valid paths that are strictly consistent with the visual scene. 
By explicitly grounding the diffusion process in the extracted scene tokens, the model establishes a robust foundation for kinematically feasible trajectory generation and spatial awareness prior to any complex semantic alignment.

\paragraph{Semantic Alignment} 
In the second stage, the action decoder is unmasked ($\lambda_{\text{action}} = 1$) and contributes to the total loss. 
This stage encourages the model to establish a latent mapping between the refined trajectories and their corresponding high-level semantic actions. 
Specifically, the diffusion decoder learns to modulate its spatial refinement based on the probabilistic belief state $\hat{\mathbf{p}}$, effectively aligning the vehicle's generated maneuvers with the high-level intention.

\section{Experiments}
\label{sec:experiments}

\subsection{Benchmark and Metrics}


To evaluate our method in a challenging instruction-following setting, we utilize the LangAuto benchmark~\cite{shao2024lmdrive} within the CARLA simulator~\cite{dosovitskiy2017carla}. 
Given that our method is specifically designed to process natural language directives rather than discrete goal points, we use LangAuto as the only existing closed-loop evaluation framework that supports such guidance. 
The benchmark challenges agents to navigate eight diverse towns across three tracks based on route length: \textit{Tiny} ($<150m$), \textit{Short} ($150-500m$), and \textit{Long} ($>500m$). 
We report standard CARLA metrics: Route Completion (RC) for the percentage of distance traversed, Infraction Score (IS) as a multiplicative safety penalty for collisions and traffic violations, and the primary composite metric, Driving Score (DS), defined as the product of RC and IS. 
Consistent with the LangAuto protocol, all reported results represent the average of three independent evaluation runs.

\subsection{Implementation Details}

\subsubsection{Agent and Controller Optimization}
Beyond standard baseline evaluation, we address two critical engineering bottlenecks in LMDrive, significantly improving its baseline performance and creating a highly competitive benchmark.
First, to prevent the accumulation of stale temporal noise and distributional drift over long horizons, we remove the baseline's prediction head for an instruction-based history reset and enforce a strict 40-frame sliding window during inference, matching the training distribution's receptive field. 
Second, we observed that the official LMDrive checkpoint is significantly under-calibrated for low-level lateral execution. 
Following insights from~\cite{winter2025bevdriver}, we stabilize the lateral dynamics by optimizing the PID controller coefficients, decreasing the proportional gain to 1.2 while increasing the integral and derivative gains to 0.8 and 0.35, respectively. 

\begin{table*}[ht!]
  \caption{LangAuto Performance. Best results highlighted in bold.}
  \label{tab:prompt_config}
  \centering
  \begin{tabular}{l c ccc ccc ccc | ccc}
    \toprule
    \multicolumn{1}{c}{Method} & Venue & 
      \multicolumn{3}{c}{Tiny} & 
      \multicolumn{3}{c}{Short} & 
      \multicolumn{3}{c}{Long} & 
      \multicolumn{3}{|c}{Mean} \\
    \cmidrule(lr){3-5}\cmidrule(lr){6-8}\cmidrule(lr){9-11}\cmidrule(l){12-14}
    & & DS $\uparrow$ & RC $\uparrow$ & IS $\uparrow$ & DS $\uparrow$ & RC $\uparrow$ & IS $\uparrow$ & DS $\uparrow$ & RC $\uparrow$ & IS $\uparrow$ & DS $\uparrow$ & RC $\uparrow$ & IS $\uparrow$ \\
    \midrule
    LMDrive~\cite{shao2024lmdrive} (Reported) & CVPR'24 & 66.5 & 77.9 & 0.85 & 50.6 & 60.0 & 0.84 & 36.2 & 46.6 & 0.81 & 51.1 & 61.5 & 0.83 \\
    LMDrive~\cite{shao2024lmdrive} (Checkpoint) & CVPR'24 & 60.7 & 71.0 & 0.82 & 41.3 & 57.0 & 0.79 & 26.8 & 36.5 & 0.77 & 42.9 & 54.8 & 0.79 \\
    AD-H~\cite{zhang2024ad} & arXiv'24 & 77.5 & 85.1 & 0.91 & 56.1 & 68.0 & 0.78 & 44.0 & 53.2 & 0.83 & 59.2 & 68.8 & 0.84 \\
    BEVDriver~\cite{winter2025bevdriver} & IROS'25 & 70.2 & 81.3 & 0.87 & 66.7 & 77.8 & 0.87 & 48.9 & \textbf{59.7} & 0.82 & 61.9 & 72.9 & 0.85 \\
    SToRM~\cite{kim2026storm} & arXiv'26 & 78.8 & 86.9 & 0.92 & 64.5 & 74.7 & 0.88 & 44.2 & 56.8 & 0.82 & 62.5 & 72.8 & 0.87 \\
    VLDrive~\cite{zhang2025vldrive} & ICCV'25 & 81.9 & 85.5 & 0.94 & 67.4 & 78.1 & 0.85 & 43.8 & 54.5 & 0.84 & 64.4 & 72.7 & 0.88 \\
    AdaDrive~\cite{zhang2025adadrive} & ICCV'25 & 80.9 & \textbf{87.6} & 0.90 & 70.6 & \textbf{85.3} & 0.81 & 42.9 & 53.4 & 0.82 & 64.8 & \textbf{75.4} & 0.84 \\
    \midrule
    \rowcolor{gray!15} 
    \textbf{LAD-Drive (Ours)}        &     & \textbf{83.5} & 87.0 & \textbf{0.95} & \textbf{71.3} & 78.1 & \textbf{0.89} & \textbf{49.8} & 58.3 & \textbf{0.86} & \textbf{68.2} & 74.5 & \textbf{0.90} \\
    \bottomrule
  \end{tabular}
\end{table*}

\begin{table*}[ht!]
  \caption{Detailed Infraction and Efficiency Analysis, including mean total model latency [ms] and decoder parameter count [M].}
  \label{tab:safety_metrics}
  \centering
  \begin{tabular}{l ccc ccccccccc | cc}
    \toprule
    \multicolumn{1}{c}{Method} &
      \multicolumn{1}{c}{DS $\uparrow$} & \multicolumn{1}{c}{RC $\uparrow$} & \multicolumn{1}{c}{IS $\uparrow$} &
      \multicolumn{1}{c}{CP $\downarrow$} & \multicolumn{1}{c}{CV $\downarrow$} & \multicolumn{1}{c}{CL $\downarrow$} &
      \multicolumn{1}{c}{RL $\downarrow$} & \multicolumn{1}{c}{SS $\downarrow$} & \multicolumn{1}{c}{Off $\downarrow$} &
      \multicolumn{1}{c}{RD $\downarrow$} & \multicolumn{1}{c}{TO $\downarrow$} & \multicolumn{1}{c}{AB $\downarrow$} & \multicolumn{1}{|c}{DP $\downarrow$} & \multicolumn{1}{c}{Latency $\downarrow$} \\
    \midrule
    LMDrive & 42.9 & 54.8 & 0.80 & 0.08 & 2.83 & 4.28 & 2.31 & \textbf{0.00} & 3.79 & 11.95 & 2.93 & \textbf{1.63} & 33.61 & \textbf{40.15} \\
    \rowcolor{gray!15} %
    \textbf{LAD-Drive (Ours)}        & \textbf{68.2} & \textbf{74.5} & \textbf{0.90} & \textbf{0.02} & \textbf{0.67} & \textbf{1.20} & \textbf{1.60} & 0.11 & \textbf{1.10} & \textbf{2.31} & \textbf{2.68} & 2.48 & \textbf{32.13} & 47.04\\
    \bottomrule
  \end{tabular}
  
  \vspace{1.5mm}
  \raggedright
  \footnotesize{CP: Collision Pedestrians, CV: Collision Vehicles, CL: Collision Layout, RL: Red Light, SS: Stop Sign, Off: Off-Road, RD: Route Deviation, TO:~Timeout, AB: Agent Blocked, DP: Decoder Parameter Count, Latency measured on a single NVIDIA L40s GPU.}
  \vspace{-0.3cm}
\end{table*}

\subsubsection{Training Configuration}
We initialize the AD-MLLM using the official LMDrive checkpoint~\cite{shao2024lmdrive}, freezing the LLaVA-v1.5~\cite{liu2024llava} backbone and vision encoder to preserve pre-trained semantic knowledge while only fine-tuning the Q-Former and training the proposed action and diffusion decoders from scratch. Replacing the standard regression head, our diffusion decoder employs an efficient truncated denoising strategy initialized from $N_a=20$ clustered anchors with only two denoising steps for both training and inference. 
The model is trained on a subset of the GraphPilot dataset~\cite{schmidt2025graphpilot}. We use the raw sensor data and navigation instructions but deliberately omit scene-graph annotations to ensure the training protocol remains comparable to the original LMDrive baseline.
During inference, the highest-scoring trajectory is selected to guide longitudinal and lateral PID controllers in generating discrete control commands.
We employ eight NVIDIA L40s GPUs with a global batch size of 16 and a weight decay of 0.1, following a linear warmup cosine learning rate schedule that starts at $5 \times 10^{-7}$, peaks at $2 \times 10^{-5}$, and decays to $1 \times 10^{-6}$.
The training spans a total of six epochs, divided equally into the two proposed stages, and requires approximately seven hours to complete.

\subsection{Results}

\begin{figure*}[ht!]
\centering
\includegraphics[width=\linewidth]{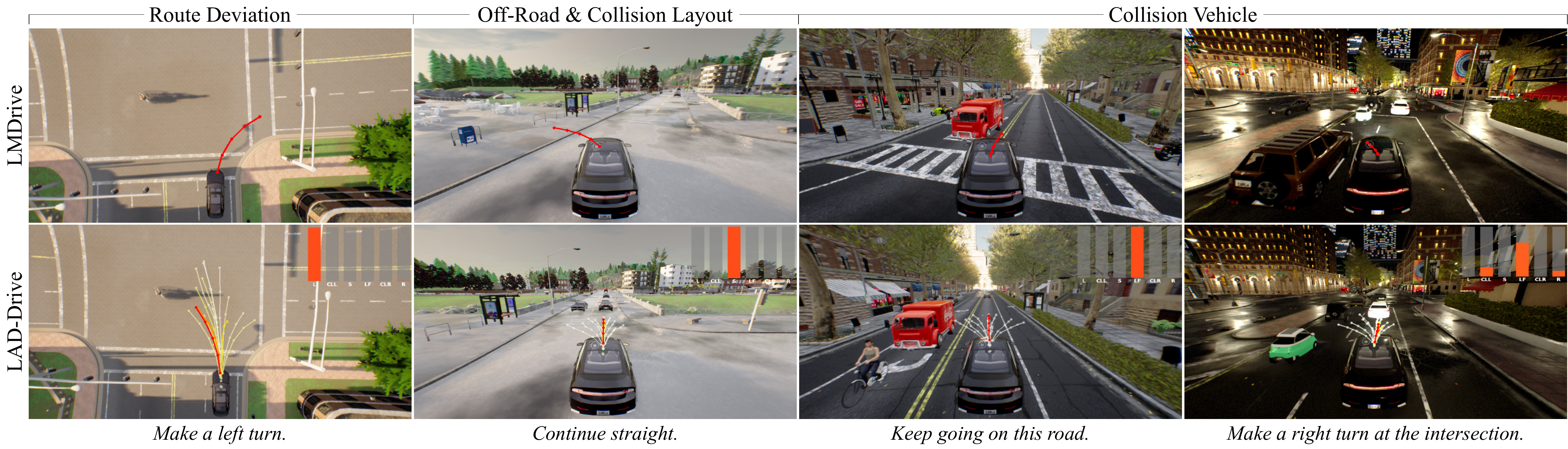}
\caption{Qualitative comparison between LMDrive (top row) and LAD-Drive (bottom row). For LAD-Drive, the generated multimodal trajectories are visualized using a heat map color scheme, with paths ranging from white (low confidence) to red (high confidence). The predicted meta-action distribution from the action decoder is displayed in the top right of each LAD-Drive panel (L: Left, CLL: Change Lane Left, S: Straight, LF: Lane Follow, CLR: Change Lane Right, R: Right). Column 1 (Route Deviation): Despite the instruction to make a left turn, LMDrive turns right. LAD-Drive executes the correct intention while safely avoiding a cyclist. Column 2 (Off-Road \& Collision Layout): LMDrive fails to interpret the road layout and instruction, driving off-road into a fence, whereas LAD-Drive successfully continues straight. Columns 3 \& 4 (Collision Vehicle): In the third scenario, LMDrive collides head-on with a truck, while LAD-Drive safely navigates the lane. In the fourth scenario, LMDrive initiates an unsafe left lane change into traffic, ignoring the right-turn navigation instruction. In contrast, LAD-Drive effectively handles dynamic agents and remains safely on the road, demonstrating robustness even when the action decoder exhibits slight uncertainty.}\label{fig:qualitative_results}
\end{figure*}

We present the comparison of our method against the baseline LMDrive~\cite{shao2024lmdrive} and current state-of-the-art methods in Tab.~\ref{tab:prompt_config}. 
LAD-Drive achieves a new state-of-the-art performance with a mean DS of 68.2, outperforming the LMDrive baseline (42.9) by a substantial margin of +59\% and surpassing recent approaches such as AdaDrive (64.8) and VLDrive (64.4).

We attribute the substantial improvements in RC compared to the baseline to the structural disentanglement of intention and execution.
The baseline's significantly lower RC values, which are particularly evident in both \textit{Short} and \textit{Long} benchmarks, indicate fundamental limitations in instruction-following capability, which frequently result in route deviations and consequent early episode termination. 
In contrast, our method mitigates these issues, maintaining robust adherence to navigation goals over extended horizons. 
Furthermore, even when compared with state-of-the-art methods that achieve comparable RC levels, such as BEVDriver, VLDrive, and AdaDrive, LAD-Drive consistently achieves a higher IS. 
This distinction validates the effectiveness of our disentangled architecture. By explicitly conditioning the diffusion process on the predicted meta-action distribution, our probabilistic action conditioning effectively regularizes the driving behavior, ensuring that driving progress does not come at the expense of safety.

\subsection{Ablations}

\begin{table*}[t!]
  \centering
  
  \begin{minipage}{0.57\linewidth}
    \centering
    \caption{Ablation Study on Model Components.}
    \label{tab:ablation_study}
    \begin{tabular}{c c c c c | c c c}
      \toprule
      Traj. Head & FT \& Opt. & Lat. Act. & Long. Act. & Ego & DS $\uparrow$ & RC $\uparrow$ & IS $\uparrow$ \\
      \midrule
      Baseline  & & & & & 42.9 & 54.8 & 0.79 \\
      Baseline  & \checkmark & & & & 60.6 & 66.1 & \textbf{0.91} \\
      Diffusion & \checkmark & & & & 59.4 & 66.0 & 0.89 \\
      Diffusion & \checkmark & \checkmark & & & 60.0 & 68.1 & 0.88 \\
      Diffusion & \checkmark & & & \checkmark & 64.8 & 72.2 & 0.88 \\
      \rowcolor{gray!15} %
      Diffusion & \checkmark & \checkmark & & \checkmark & \textbf{68.2} & \textbf{74.5} & 0.90 \\
      Diffusion & \checkmark & \checkmark & \checkmark & \checkmark & 65.9 & 72.4 & 0.90 \\
      \bottomrule
    \end{tabular}
    
    \vspace{1.5mm}
    \raggedright
    \footnotesize{FT \& Opt.: Fine-Tuning and Controller Optimization, Lat. Act.: Lateral Action, Long. Act.: Longitudinal Action, Ego: Ego-Status.}
  \end{minipage}%
  \hfill
  \begin{minipage}{0.35\linewidth}
    \centering
    \caption{Ablation Study on Latent Dimension of Diffusion Decoder.}
    \label{tab:latent_ablation_study}
    \begin{tabular}{c | c c c}
      \toprule
      Lat. Dim. & DS $\uparrow$ & RC $\uparrow$ & IS $\uparrow$ \\
      \midrule
      128  & 63.6 & 72.3 & 0.87 \\
      256  & 66.9 & 73.9 & 0.90 \\
      \rowcolor{gray!15} %
      512  & \textbf{68.2} & \textbf{74.5} & 0.90 \\
      1024 & 63.2 & 68.1 & 0.91 \\
      4096 & 48.5 & 52.4 & \textbf{0.92} \\
      \bottomrule
    \end{tabular}

    \vspace{1.5mm}
    \footnotesize{Lat. Dim.: Latent Dimension.}
  \end{minipage}
  \vspace{-2.5mm}
\end{table*}

To further investigate the source of these improvements, Tab.~\ref{tab:safety_metrics} details the infraction types. 
Compared to the official LMDrive checkpoint, our method drastically reduces collision rates with dynamic agents. Collision with vehicles (CV) drops by 76\% (2.83 to 0.67) and collision with pedestrians (CP) decreases by 80\% (0.08 to 0.02). A safety improvement is visually depicted in the right columns of Fig.~\ref{fig:qualitative_results}. 
Equally important is the improvement in spatial grounding. Violations related to physical control, such as collision with layout (CL) and off-road (Off) events, are reduced by approximately 72\% and 71\%, respectively. 
As illustrated in the middle column of Fig.~\ref{fig:qualitative_results}, this confirms that our diffusion decoder generates trajectories that are not only semantically consistent but also physically compliant with the road boundaries.
The Route Deviation (RD) metric improves significantly, decreasing from 11.95 to 2.31. 
As demonstrated in the left column of Fig.~\ref{fig:qualitative_results}, this reduction can be primarily attributed to our effective probabilistic meta-action conditioning, which ensures that the diffusion decoder remains strictly aligned with the intended navigation goal. 
In contrast, standard regression-based baselines are often computationally overloaded, as they attempt to resolve high-level semantic intentions and low-level spatial control simultaneously within a single head. 
This dual burden frequently results in erroneous lateral decisions, such as incorrect turns at intersections, which cause the agent to terminate episodes prematurely. 
By disentangling these tasks, LAD-Drive effectively regularizes the trajectory generation process to prevent navigational failures. 
This architectural refinement achieves a significant performance leap while actually reducing the decoder footprint by 1.48M parameters compared to the baseline.
Although the two-step denoising process introduces a modest increase in inference overhead (+6.89ms), the model maintains a highly competitive total real-time latency of 47.04ms.

To validate the contribution of each architectural component, we conduct a systematic ablation study presented in Tab.~\ref{tab:ablation_study}. 
Initial results demonstrate that fine-tuning the base model and optimizing the agent (row 2) yields a substantial performance leap over the original checkpoint (DS 42.9 to 60.6). 
This establishes a strong regression baseline, ensuring that subsequent gains are attributable to architectural innovations rather than training recipe or parameter improvements. 
Simply substituting the regression head with a diffusion decoder (row 3) results in a slight performance degradation (DS 60.6 to 59.4). 
This confirms that while generative models offer multimodal capacity, they are inherently unstable without sufficient state guidance, often producing diverse but physically inconsistent trajectories.
We observe a critical trade-off when conditioning mechanisms are isolated. 
Integrating only the lateral action decoder (row 4) provides semantic guidance on what to do, yielding a marginal overall improvement (DS 60.0). 
However, the lack of physical state awareness prevents the model from executing these intentions consistently over time. 
In contrast, relying solely on ego-status grounding (row 5) significantly improves physical stability (DS 64.8) by ensuring temporal consistency, yet without explicit action predictions, the model struggles to resolve navigational ambiguities in complex scenarios. 
The combined configuration (row 6) achieves the highest performance (DS 68.2), validating the necessary synergy. High-level lateral action guidance resolves semantic ambiguity while low-level ego-grounding ensures physical feasibility. 
Finally, introducing a longitudinal action head (row 7) degrades performance (DS 65.9). 
This empirical result corroborates our hypothesis that as longitudinal dynamics are largely governed by the vehicle's ego-status and immediate spatial constraints, forcing the diffusion process to also condition on a high-level longitudinal command introduces redundant semantic noise that disrupts the refinement process.

Tab.~\ref{tab:latent_ablation_study} details the impact of the diffusion decoder's latent dimension size. 
We observe a clear performance peak at $d=512$, which optimally balances feature compression and representational capacity to achieve the highest performance (DS 68.2). 
Constraining the bottleneck too aggressively ($d=128$) restricts the model's ability to capture complex scene dynamics, whereas expanding it beyond 512 degrades navigational accuracy. Most notably, bypassing the bottleneck to directly couple the raw 4096-dimensional LLM embeddings results in a severe performance collapse (DS 48.5).
This empirical drop explicitly confirms our structural hypothesis that in the absence of a semantic filter, the model suffers from a severe representation mismatch and dimensionality overload.\looseness=-1

\section{Conclusion}

We introduced LAD-Drive to resolve the modality gap between high-level language reasoning and continuous trajectory planning. 
LAD-Drive employs an action decoder to infer a probabilistic meta-action distribution, structurally disentangling semantic intention from spatial execution.
This forms an explicit belief state that preserves critical navigational uncertainty. 
Fused with the vehicle's kinematic ego-status, this distribution provides robust probabilistic conditioning for an action-aware diffusion decoder. 
Coupled with a structural feature bottleneck that filters high-dimensional semantic noise from the LLM, our framework efficiently translates abstract reasoning into safe, executable trajectories. 
Extensive evaluations on the LangAuto benchmark validate our approach. 
LAD-Drive achieves state-of-the-art performance, outperforming the baseline by 59\% in Driving Score. At the same time, this progress is accompanied by substantial improvements in safety and instruction following, reducing vehicle and layout collisions by 76\% and 72\%, respectively, while decreasing route deviations by 81\%.






\bibliographystyle{IEEEtran}
\bibliography{bibliography}

\addtolength{\textheight}{-12cm}   

\end{document}